\pdfoutput=1
\documentclass[11pt]{article}

\usepackage[preprint]{acl}
\usepackage{lmodern}

\usepackage{times}
\usepackage{latexsym}

\usepackage[T1]{fontenc}

\usepackage[utf8]{inputenc}

\usepackage{microtype}
\usepackage{amssymb}
\usepackage{mathtools} 
\newcommand{\eqdef}{\mathrel{\coloneqq}}
\usepackage{algorithm}
\usepackage{algorithmic}

\usepackage{multirow}
\usepackage{threeparttable}

\usepackage{float}
\usepackage{graphicx}
\usepackage{subcaption}

\usepackage{multirow}
\usepackage{amsmath}
\usepackage{amssymb}
\usepackage{scalerel}
\usepackage[inline]{enumitem}
\usepackage{listings}
\usepackage{varwidth}
\usepackage[export]{adjustbox}
\usepackage{cleveref}
\usepackage{stmaryrd}
\usepackage{bbm}

\usepackage{inconsolata}

\usepackage{graphicx}

\title{Direct Preference Knowledge Distillation for Large Language Models}

\author{
 \textbf{Yixing Li\textsuperscript{1}}\thanks{~Contribution during internship at Microsoft Research.},
 \textbf{Yuxian Gu\textsuperscript{2}}\footnotemark[1],
 \textbf{Li Dong\textsuperscript{3}},
 \textbf{Dequan Wang\textsuperscript{1}},
 \textbf{Yu Cheng\textsuperscript{4}},
 \textbf{Furu Wei \textsuperscript{3}}
\\
 \textsuperscript{1}Shanghai Jiao Tong University~~
 \textsuperscript{2}Tsinghua University
 \\
 \textsuperscript{3}Microsoft Research~~
 \textsuperscript{4}The Chinese University of Hong Kong
\\
\texttt{\{lyxing0, dequanwang\}@sjtu.edu.cn}~~
\texttt{guyx21@mails.tsinghua.edu.cn}
\\
\texttt{\{lidong1,fuwei\}@microsoft.com}~~
\texttt{chengyu@cse.cuhk.edu.hk}
}

\begin{document}
\maketitle

\begin{abstract}

In the field of large language models (LLMs), Knowledge Distillation (KD) is a critical technique for transferring capabilities from teacher models to student models. However, existing KD methods face limitations and challenges in the distillation of LLMs, including efficiency and insufficient measurement capabilities of traditional Kullback-Leibler (KL) divergence. 
 It is shown that LLMs can serve as an implicit reward function, which we define as a supplement to KL divergence.
In this work, we propose Direct Preference Knowledge Distillation (DPKD) for LLMs. 
DPKD utilizes distribution divergence to represent the preference loss and the implicit reward function. 
We re-formulate KD of LLMs into two stages: first optimizing an objective consisting of implicit reward and reverse KL divergence and then improving the preference probability of teacher outputs over student outputs. We conducted experiments on various datasets with LLM parameters ranging from 120M to 13B and demonstrate the broad applicability and effectiveness of our DPKD approach. Meanwhile, we prove the value and effectiveness of the introduced implicit reward and output preference in KD through experiments and theoretical analysis. The DPKD method outperforms the baseline method in both output response precision and exact match percentage. 
\end{abstract}

\section{Introduction}



In the era of Large Language Models (LLMs), a series of models and techniques have demonstrated great capabilities \cite{zhang2022opt, chowdhery2023palm}
, and their powerful capabilities are usually ascribed to the increase in the size of the training data and the scale \cite{kaplan2020scaling,anil2023palm}. However, the increase in the size of large models also brings expensive computing costs and difficulties \cite{hoffmann2022training}. How to maintain performance and conduct efficient training while increasing the scale of language models has become an important issue.
Knowledge distillation (KD;~\citealp{hinton2015distilling})  trains a relatively compact student model by simulating the output distribution and behavior of the teacher model, which is an effective method under limited computing resources. 


Recent work has recognized the shortcomings of KL divergence in traditional KD in which of LLMs \cite{Gu2023KnowledgeDO}, and explored other forms of KL divergence like reverse KL divergence.
Work on KL divergence \cite{wu2024rethinking} proposed different KLD metrics for different stages of their respective shortcomings, and proposed improvement methods to expand KL divergence into flexible distance form. 
In fact, KL divergence is insufficient under the condition of a stronger teacher model \cite{huang2022knowledge,cho2019efficacy,son2021densely}, requiring the introduction of additional objectives and innovative knowledge distillation procedures.
We start from another perspective and consider a novel optimization objective to compensate for the above problems while maintaining high efficiency.

In this work, we propose Direct Preference Knowledge Distillation (DPKD) for the knowledge distillation of LLMs.  It is shown that LLMs can serve as an implicit reward function \cite{yuan2024self,rafailov2024direct}. Due to the deficiency of KL divergence \cite{Gu2023KnowledgeDO, huang2022knowledge}, we define implicit reward function as a supplement to the KL distance. 
We re-formulate the KD of white-box LLMs as follows, first maximize the optimization function consisting of implicit reward and reverse KL divergence, and then improve the preference probability of teacher outputs over student outputs. These settings compensate for the shortcomings of KL divergence while preserving high training efficiency.  We derive the training objective to obtain a concise final form, and demonstrate the significance of the reward and preference form we introduced in KD through theoretical derivation.


Finally, we conducted experiments on our DPKD method in the instruction tuning task. We utilized various families of LLMs including GPT-2 \cite{Radford2019LanguageMA} and OPT \cite{Zhang2022OPTOP} with parameter sizes ranging from 120M to 13B and multiple datasets 
to validate our method. We evaluate the results with Rouge-L \cite{lin-2004-rouge}  
metric. Our results show that the DPKD method outperforms the baseline and has advantages over a wide range of generation lengths. Additionally, we conducted experiments on the implicit reward function in the reformulation of distillation to verify its effectiveness, and experiments on the forms of preference to show the potential of preference modeling in KD and provide ideas for subsequent research. Our contributions are thus three-fold:
\begin{itemize}[leftmargin=*]
\item We propose a novel framework called Direct Preference Knowledge Distillation (DPKD), which provides a new perspective besides designing different KL divergences for knowledge distillation of large models.
\item We compare DPKD to five baseline
methods across two commonly used LLMs (GPT-2 and OPT) and three
datasets. We also provide a detailed derivation and theoretical analysis to demonstrate the effectiveness of the reward and preference model introduced in this paper.
\item We provide additional experiments on other preference objectives and observations on implicit reward. We illustrate the effectiveness and potential of our reformulation of the KD process, providing inspiration for subsequent work.
\end{itemize}

\section{Methods}

\subsection{Preliminaries}

\subsubsection{Sequence-Level Knowledge Distillation}

Sequence-level KD is often formulated as an optimization problem. Given a fixed teacher model with output distribution $p$ and a student model with output distribution $q_\theta$, the optimization goal is to minimize the distribution distance between  $p$ and  $q_\theta$. The distance is measured by Kullback-Leibler divergence (KLD). In the case of KD for LLMs, forward-KLD and reverse-KLD (rKLD) are the most studied, respective defined as $\mbox{forward-KLD =} \mathbb{E}_{x\sim p_x, y\sim p}\log \frac{p(y \mid x)}{q_{\theta}(y \mid x)}$ and $\mbox{reverse-KLD =} \mathbb{E}_{x\sim p_x, y\sim q_\theta}\log \frac{q_{\theta}(y \mid x)}{p(y \mid x)}$. The reverse KL divergence is more suitable for KD while the distribution of LLMs is more complicated.

\subsubsection{Preference Model}

Given two comparable objects or events, a common model for comparing the probabilities of their selection is the Bradley-Terry (BT;~\citealp{Bradley1952RankAO}) model. We consider output $y_1$ and $y_2$, and the reward function that measures the gain of choosing an output is denoted $r(y)$. BT model can be used to measure the probability that we choose $y_1$ instead of $y_2$:

\begin{small}
\begin{equation}
\begin{aligned}
  \label{eq:example}
 Pr\left(y_1 \succ y_2 \right) &=\frac{\exp \left(r\left(y_1\right)\right)}{\exp \left(r\left( y_1\right)\right)+\exp \left(r\left( y_2\right)\right)}\\
 &= \sigma\left(r\left( y_1\right)- r\left(y_2\right) \right),
 \end{aligned}
\end{equation}
\end{small}
where $\sigma$ is the function $\sigma(x) \eqdef \frac{1}{1 + \exp(-x)}$.

\subsection{DPKD: Direct Preference Knowledge Distillation}
\label{DPKD-method}


In the context of knowledge distillation of LLMs, the difference in distribution measured by KL divergence is  often regarded as the only criterion for the distillation target. But KL divergence is insufficient. 
KL divergence measures the difference between two distributions, but it is not really a distance because it is asymmetric \cite{manning1999foundations}. Researchers have improved the KL divergence by combining two symmetrical items (Jensen-Shannon divergence \cite{nielsen2021variational}). In the field of KD, researchers have explored the inadequacy of KL divergence in KD \cite{cho2019efficacy,mirzadeh2020improved}. Some works \cite{Gu2023KnowledgeDO,wu2024rethinking} hope to redesign KL divergence, and some works\cite{huang2022knowledge} add new objectives besides KL divergence. 
We conducted a toy experiment to further illustrate the inadequacy of KLD in Figure \ref{fig:toyexperiment} and Section \ref{sec:Reward Observation}. 

Based on the above considerations, we denote the implicit reward  $r_p \left(  \mathbf{y} | \mathbf{x}\right)$ as a supplement to the KL divergence in the optimization objective. We formulate the  optimization goal as:

\begin{small}
\begin{equation}
\max_{\theta} \mathbb{E}\,[\,r_p(y|x) - \beta \,  \text{KLD}\,\,(q_{\theta}(y|x)) \| p(y|x) )\,] . \label{eq2}
\end{equation}
\end{small}

We can get the optimal solution of Equation~\ref{eq2} through the following steps. Firstly transform Equation~\ref{eq2} as follows:

\begin{small}
\begin{equation}
\begin{aligned}
 & \min _\theta \mathbb{E}\left[\log \frac{q_{\theta}(y \mid x)}{p(y \mid x)}-\frac{1}{\beta} r(x, y)\right] \\
 =&\min _\theta \mathbb{E}\left[\log \frac{q_{\theta}(y \mid x)}{q^{*}(y \mid x)}-\log Z(x)\right],\label{eq3}
\end{aligned}
\end{equation}
\end{small}
where $q^{*}(y | x) \eqdef \frac{1}{Z(x)} p(y | x) \exp \left(\frac{1}{\beta} r(x, y)\right)$, and $Z\left( x \right)$ is the scaling function of the distribution, which is required to be independent of $\theta$ and  $y$. It is defined as $Z(x) \eqdef \sum_y p(y | x) \exp \left(\frac{1}{\beta} r(x, y)\right)$. Following the work, we can derive the optimal solution of Equation~\ref{eq3}:
\begin{small}
\begin{equation}
    q_{\theta}(y \mid x) = q^*(y \mid x) . \label{eq4}
\end{equation}
\end{small}
From Equation~\ref{eq4} we can obtain that:

\begin{small}
\begin{equation}
    r^*(x, y)=\beta \log \frac{q^*(y \mid x)}{p(y \mid x)}+\beta \log Z(x) . \label{eq5}
\end{equation}
\end{small}

Given the same prompt x, the outputs of the student model and the teacher model are denoted as $y_s$ and $y_t$ respectively. The purpose of KD is to fit the distribution of the student model to teacher model, which can also be understood as we expect the student model to have a greater probability of outputting results similar to those of the teacher model. From the BT model, we can obtain  the following probability:

\begin{small}
\begin{equation}
\begin{aligned}
    p&^* \left(y_t \succ y_s \mid x\right) \\ 
    & = \sigma\left(\beta \log \frac{q_{\theta}\left(y_t \mid x\right)}{p\left(y_t \mid x\right)}-\beta \log \frac{q_{\theta}\left(y_s \mid x\right)}{p\left(y_s \mid x\right)}\right). \label{3-11}
\end{aligned}
\end{equation}
\end{small}
The complete derivation of the above procedure is in the Appendix \ref{sec:appendix dpkd objective}. Since our goal is to maximize the probability of model outputing $y_t$ rather than $y_s$, which is $\max _\theta \,\, \mathbb{E} \log \, p^*\left(y_t \succ y_s \mid x\right)$. It is formulated as:

\begin{small}
\begin{equation}
\min_\theta -\mathbb{E}\left[\log \sigma\left(\beta \log \frac{q_{\theta}\left(y_t \right)}{p\left(y_t \right)}-\beta \log \frac{q_{\theta}\left(y_s \right)}{p\left(y_s \right)}\right)\right] . 
\end{equation}
\end{small}

where $q_{\theta}\left(y_t \right)$ denotes $q_{\theta}\left(y_t \mid x\right)$ for short, and same for $p$ and $y_s$.

\subsubsection{Learning Objective}\label{sec:Learning Objective}

Following the framework derivation of DPKD, the loss is defined as:

\begin{small}
\begin{equation}
    \mathcal{L} = -\mathbb{E}\left[\log \sigma\left(\beta \log \frac{q_{\theta}\left(y_t \right)}{p\left(y_t \right)}-\beta \log \frac{q_{\theta}\left(y_s \right)}{p\left(y_s \right)}\right)\right].
\end{equation}\label{DPKD-loss}
\end{small}

It is shown by works \cite{meng2024simpo, Gu2023KnowledgeDO} that models tend to introduce bias and produce short responses without length normalization. We add the length normalization factor to the distillation loss as follows:

\begin{small}
\begin{equation}
\mathcal{L} = -\mathbb{E}\left[\log \sigma\left(\frac{\beta}{\lvert y_t \rvert} \log \frac{q_{\theta}\left(y_t \right)}{p\left(y_t \right)}-\frac{\beta}{\lvert y_s \rvert} \log \frac{q_{\theta}\left(y_s \right)}{p\left(y_s \right)}\right)\right].
\end{equation}
\end{small}

Following work \cite{ouyang2022training, Gu2023KnowledgeDO}, we add the language modeling loss in order to preserve performance on
canonical NLP benchmarks. The complete algorithm process is shown in Algorithm~\ref{alg:1}.

\begin{algorithm}[t]
\caption{Direct Preference Knowledge Distillation (DPKD)}
\label{alg:1}
\begin{algorithmic}[1] 
\REQUIRE ~~\\ 
    Instruction tuning datasets $D$; \\
    Pre-training corpus $D_p$; \\
    Fine-tuned teacher model with distribution $p$;\\
    Initialized student model conducted SFT on $D_p$ with output distribution $q_\theta$; \\
\ENSURE ~~\\ 
    Student model parameter conducted distillation training;
       \FOR{epoch in epochs}
            \FOR{batch in datasets $D$ and $D_p$}
            \STATE Compute responses from teacher and student model and obtain $y_t$ and $y_s$;
            \STATE Compute four log items $\log p(y_t)$, $\log p(y_s)$,  $\log q_\theta(y_t)$ and $\log q_\theta(y_s)$; 
            \STATE Compute distill loss $ \mathcal{L}_{kd}  =  -\sum_{\left(x\sim \mathcal{D} \right) }$;
 \\ $ \log \sigma \left(\frac{\beta}{\lvert y_t \rvert} \log \frac{q_{\theta}\left(y_t \right)}{p\left(y_t \right)}-\frac{\beta}{\lvert y_s \rvert} \log \frac{q_{\theta}\left(y_s \right)}{p\left(y_s \right)}\right)$;
            \STATE Compute language modeling loss $ \mathcal{L}_{pt}  =  -\sum_{d\sim \mathcal{D}_{p} } \log q_\theta (d) $;
            \STATE Compute loss $ \mathcal{L} = \mathcal{L}_{kd} + \lambda \cdot \mathcal{L}_{pt}$ and gradient $\nabla_\theta \mathcal{L}$;
            \STATE Gradient Update $\theta^{(t+1)}=\theta^{(t)}-\alpha \nabla_\theta \mathcal{L}$;
            \ENDFOR
        \ENDFOR
    
\RETURN Student model parameter $\theta$. 
\end{algorithmic}
\end{algorithm}


\section{Analysis}

\subsection{Gradient Derivation}\label{sec:Gradient Derivation}
Based on our target definition, we can derive the gradient of DPKD through the basic chain rule and perform analysis similar to work \cite{rafailov2024direct}. From Equation~\ref{DPKD-loss}, we can perform variable substitution and define $u \eqdef \beta \log \frac{q_\theta\left(y_s \mid x\right)}{p\left(y_s \mid x\right)}-\beta \log \frac{q_\theta\left(y_t \mid x\right)}{p\left(y_t \mid x\right)}$, and the gradient of the loss can be expressed as

\begin{small}
\begin{equation}
\begin{aligned}
   \nabla_\theta \mathcal{L} & = -\nabla_\theta \mathbb{E}_{\left(x, y_t, y_s\right) \sim \mathcal{D}}\left[\log \sigma\left(u\right)\right]\\
   &= -\mathbb{E}_{\left(x, y_t, y_s\right) \sim \mathcal{D}}\left[\frac{\sigma^{\prime}(u)}{\sigma(u)} \nabla_\theta(u)\right]. \label{dpkd-loss-dif1}
\end{aligned}
\end{equation}
\end{small}

We can derive the complete form of gradient:

\begin{small}
\begin{equation}
\begin{aligned}
\nabla_\theta \mathcal{L} =  & -\mathbb{E}\left[\beta  \sigma\left(\beta \log \frac{q_\theta\left(y_t \right)}{p\left(y_t \right)}-\beta \log \frac{q_\theta\left(y_s \right)}{p\left(y_s \right)}\right) \right. \\
& \left. \left[\nabla_\theta \log q_\theta\left(y_t \right)-\nabla_\theta \log q_\theta\left(y_s \right)\right]\right].\label{dpkd-loss-diff-final}
\end{aligned}
\end{equation}
\end{small}

The full derivation of the gradient is in the Appendix \ref{sec:appendix dpkd graadient}.

\subsection{Optimizing DPKD Is Optimizing The Q Function}


Following work \cite{rafailov2024r}, we conduct a theoretical analysis of DPKD and relate the reward function in KD we introduced to the Q function..We consider the sequence generation task, and the data set $\mathcal{D}=\{(\mathbf{x},\mathbf{y})\}^{N}$, where the prompt is denoted as $\mathbf{x} = \{x_0, \cdots, x_{l-1}\}$. We denote the vocabulary as $\mathbf{V}$, each word $x$ is included in the vocabulary $\mathbf{V}$. Given input prompt $\mathbf{x}$, the model output is $\mathbf{y} = \{y_0, \cdots, y_{m-1}\}$, where $m$ is the maximum generated length. At each time step, the generated $y_t$ is conditionally generated based on the generated sequence $\{ \mathbf{x}, \mathbf{y_{t-1}}\}$.

From another perspective of Markov Decision Process (MDP), the process of sequence generation by the model is regarded as the generation process of a Markov decision chain. The state is denoted as $s_t=\{x_0,\ldots,x_{l-1},y_0,\ldots,y_{t-1}\}$, and the action is $y_t \in \mathbf{V}$, which is chosen based on generated sequence $s_t$. The transition function from the current state to the next state is the LLMs we selected. We analyze the implicit reward function of DPKD from $Q$ function, which is a perspective in reinforcement learning. The general represent of $Q$ is:

\begin{small}
\begin{equation}
    Q^{\theta}(s_{t},a_{t})=\mathbb{E}[R_{t+1}+\gamma R_{t+2}+ ... \mid s_{t},a_{t}].\label{Q-1}
\end{equation}
\end{small}

Following the MDP perspective of sequence generation  and according to the work 
, the fixed point solution of Equation~\ref{Q-1} is:

\begin{small}
\begin{equation}
    q_\theta^*(\mathbf{a}_t|\mathbf{s}_t)=e^{(Q^*(\mathbf{s}_t,\mathbf{a}_t)-V^*(\mathbf{s}_t))/\beta}.\label{dpkd_q_2}
\end{equation}
\end{small}

Any valid $Q$ function needs to satisfy the Belmman equation, from which we can write the current step $Q^*$ function for the optimal strategy (that is, the most optimal student model parameters) and the reward function $r$ as:

\begin{small}
\begin{equation}
Q^*(\mathbf{s}_t,\mathbf{a}_t)=r(\mathbf{s}_t,\mathbf{a}_t)+\beta\log p(\mathbf{a}_t|\mathbf{s}_t)+V^*(\mathbf{s}_{t+1}),\label{dpkd_q_1}
\end{equation}
\end{small}
where $V^*(\mathbf{s})$ is defined to be zero if $s$ is the end of sequence (EOS). Following work \cite{rafailov2024r}, we can use the $Q$ function instead of $r$ and substitute it into the Bellman equation.

Now we can use the $Q$ function to re-derive the DPKD method. By transforming the Equation~\ref{dpkd_q_1} and summing over time t, and substitute Equation~\ref{dpkd_q_2} into the result, we could obtain:

\begin{small}
\begin{equation}
\begin{aligned}
\sum_{t=0}^{T-1}r(\mathbf{s}_{t},\mathbf{a}_{t}) =V^*(\mathbf{s}_0)+\sum_{t=0}^{T-1}\beta\log\frac{ q_\theta ^*(\mathbf{a}_t|\mathbf{s}_t)}{ p (\mathbf{a}_t|\mathbf{s}_t)}
\end{aligned}.\label{dpkd_q_4}
\end{equation}
\end{small}

According to the multi-step generalization of the Bradley-Terry preference model, namely the Plackett-Luce model
, and substituting the Equation \ref{dpkd_q_4} into it, we can obtain:

\begin{small}
    \begin{equation}
    p(\tau^{t}\succeq\tau^{s})= \sigma\left(\beta \log \frac{q_{\theta}\left(y_t\right)}{p\left(y_t \right)}-\beta \log \frac{q_{\theta}\left(y_s \right)}{p\left(y_s\right)}\right),
\end{equation}
\end{small}
where $\tau$ refers to the sequence trajectory generated by the model, and in the sequence generation task of the large model, it refers to the generated text. Replace $\tau^{t}$ and $\tau^{s}$ with $y_t$ and $y_s$ respectively, and the above formula is exactly the same as Equation~\ref{3-11}. Thus we get the complete process of deriving the DPKD of this work from the $Q$ function. Therefore, we can link the DPKD method of this work with the concept of $Q$ function in reinforcement learning and Markov decision chain.

\subsection{Your Language Model Is Secretly Implicit Reward Function During KD}

We implicitly optimized a reward function during the KD process by redefining the formula and objectives of KD, and our experimental results demonstrate the significance of the reward function in the KD training process. 
From the perspective of theoretical analysis, the reward function serves not only as a weight term in the gradient of our DPKD but also as a crucial link between our DPKD method and the Q function in reinforcement learning.

Furthermore, the results presented in Section \ref{sec: Deformation of Preference} indicate that various preference expressions related to reward functions yield significantly different results. Notably, some preference expressions demonstrate promising results, despite minimal parameter fine-tuning. This suggests that employing reward functions to articulate preferences in the knowledge distillation process is effective and highlights the potential for further exploration, which also offers valuable insights for future research.

\section{Experiments}

\subsection{Experiments Setup}\label{sec:Experiments Setup}

\paragraph{Tasks and Metrics.} We conduct experiments and analyses on the task of instruction tuning. The task of instruction tuning range from summarizing to completing requests, requiring the model to generate responses based on the provided instructions, prompts, and inputs. To ensure that the model produces high-quality results of sufficient length, we specifically select data with instructions and outputs exceeding 10 words in the dataset. The Rouge-L score \cite{lin-2004-rouge} is employed to assess the quality of model generation, as \citet{wang-etal-2022-super} has shown that Rouge-L is appropriate for large-scale instruction tuning evaluation.

\paragraph{Datasets.} For the datasets, we use the following datasets: (1) Dolly \cite{DatabricksBlog2023DollyV2} including $\sim$15k instruction/response generated in capability domains from the InstructGPT, (2) Self-Inst \cite{wang-etal-2023-self-instruct} consisting of 252 expert-written tasks and their instructions motivated by user-oriented applications, (3) \textsc{SuperNatural-Instructions} (S-NI;~\citealp{wang-etal-2022-super})  containing over 1.5k tasks and covering 76 distinct task types. We filter the data whose length exceeds the model processing length and select the part with response length $\left[11, \infty \right]$ as training data to keep the same settings as the validation set and test machine. 

\paragraph{Models.} For teacher models, we select GPT-2 \cite{Radford2019LanguageMA} (1.5B) and OPT \cite{Zhang2022OPTOP} (13B). The student models are GPT-2 with 120M and 340M parameters and OPT with 1.3B parameters respectively. These models have been  pre-trained on the $D_p$ datasets.


\paragraph{Baselines.} The baseline methods we selected include: (1) Supervised Fine-Tuning (\textbf{SFT}) directly fine-tune on labels of dataset; (2) \textbf{KD} \cite{hinton2015distilling} is also called word-level KD, which uses the teacher’s output distributions as supervision; (3) Sequence-level KD (\textbf{SeqKD};~\citealp{kim-rush-2016-sequence, zhou2024lima}) directly distills the student model on the data generated by the teacher model; (4) \textbf{MiniLLM} \cite{Gu2023KnowledgeDO} leverages reverse KL divergence to perform distillation on LLMs ;(5) Adaptive Kullback-Leiber (\textbf{AKL};~\citealp{wu2024rethinking}) uses a combination of forward and reverse KL with dynamic weights to perform distillation.

\subsection{Results}\label{sec:results-main}

\begin{table}[t]\small
\setlength{\tabcolsep}{2pt}
\centering
\begin{threeparttable}
\begin{tabular}{lccccc}
\hline
\textbf{Model}         & \textbf{\#Param}          & \textbf{Method} & \textbf{DollyEval} & \textbf{SelfInst} & \textbf{S-NI}     \\ \hline
\multirow{7}{*}{GPT-2} & 1.5B                       & Teacher         & 27.6               & 14.3              & 27.6                             \\ \cline{2-6} 
& \multirow{6}{*}{120M} & SFT             & 23.3               & 10.0              & 16.3                         \\
&                            & KD              & 22.8               & 10.8              & 13.4                             \\
&                            & SeqKD           & 22.7               & 10.1              & 16.4                             \\
&                            & MiniLLM         & 24.6               & 13.2              & 25.3                   \\
&                            & AKL             & 23.9              & -                & 19.2                                \\
&                            & DPKD (ours)     & \textbf{24.6}   & \textbf{13.8}  & \textbf{25.4}       \\ \hline
\multirow{6}{*}{GPT-2} & 1.5B                           & Teacher         & 27.6               & 14.3              & 27.6                              \\ \cline{2-6} 
& \multirow{5}{*}{340M} & SFT             & 25.5               & 13.0              & 25.1                              \\ 
&                                & KD              & 25.0               & 12.0              & 23.7                            \\
&                                & SeqKD           & 25.3               & 12.6              & 22.9                              \\
&                                & MiniLLM         & 25.4               & \textbf{15.6}              & 27.4                   \\
&                                & DPKD (ours)     & \textbf{26.0}   & 14.3  & \textbf{27.6 }        \\ \hline
\multirow{6}{*}{OPT} & 13B & Teacher & 29.2 & 18.4 & 30.4 \\ \cline{2-6} 
& \multirow{5}{*}{1.3B} & SFT & 26.0 & 11.4 & 23.1 \\
 &  & KD & 25.4 & 12.2 & 21.9 \\
 &  & SeqKD & 26.1 &  12.7 & 21.4 \\
 &  & MiniLLM & 26.7 & 14.8 &  28.6\\
 &  & DPKD (ours) & \textbf{27.2} & \textbf{14.9} & \textbf{28.7}  \\ \hline
\end{tabular}
  \end{threeparttable}
\caption{Experimental results with GPT-2 and GPT model families. The teacher models of the GPT-2 and OPT are 1.5B and 13B respectively. The student model ranges from 120M to 1.3B. The scores are RougeL scores, and the experiments cover the DollyEval, SelfInst, and S-NI datasets. All experiments are conducted with the same seed. AKL results are taken from \cite{wu2024rethinking} }
\label{experiments-results-rougel} 
\end{table}

We conducted experiments based on Algorithm \ref{alg:1} and the experimental settings in Section \ref{sec:Experiments Setup}, and the experimental results are presented in Table \ref{experiments-results-rougel}. In this section, we conduct a preliminary analysis of the results including the RougeL score, the GPT-4 score, and an example of the implicit reward function introduced in this article. 

From the experimental results, we could see that our method demonstrates great performance compared to the various baseline. We conduct KD experiments on datasets with different distributions and LLMs of different sizes, showing that our method is applicable to a wide range of models and data and retains good performance. In some specific datasets, the student model trained by our knowledge distillation method is even close to the performance of the teacher model.

\begin{figure}[t]
\centering
  \includegraphics[width=0.9\columnwidth]{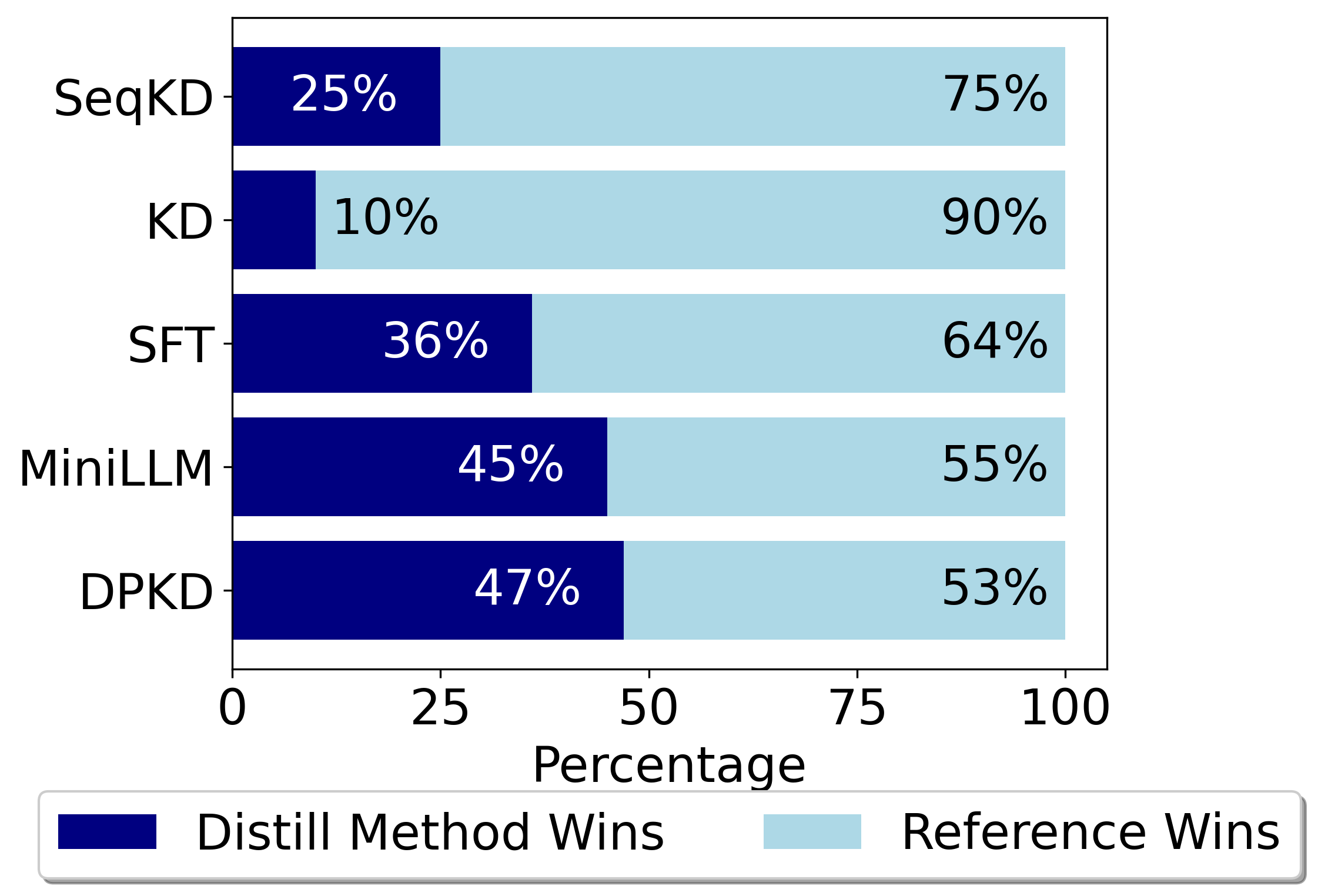} 
  \caption{GPT-4 evaluation of different methods. Our DPKD method outperforms other baselines and is closest to the reference responses.}
  \label{fig:gpt4eval}
\end{figure}

We collected the output texts of DPKD and the baseline method of GPT-2 Base, and evaluated them together with the labels of the dataset by GPT-4 to evaluate the quality of the instruction tracking generated text of each method in Figure \ref{fig:gpt4eval}. Although the output results of each method are still a certain distance away from the label, the generation results of the method proposed in this paper are closer to the label evaluations and surpass those of the baselines.



\subsection{Reward Observation}\label{sec:Reward Observation}

\begin{figure}[t]
\centering
\includegraphics[width=0.8\columnwidth]{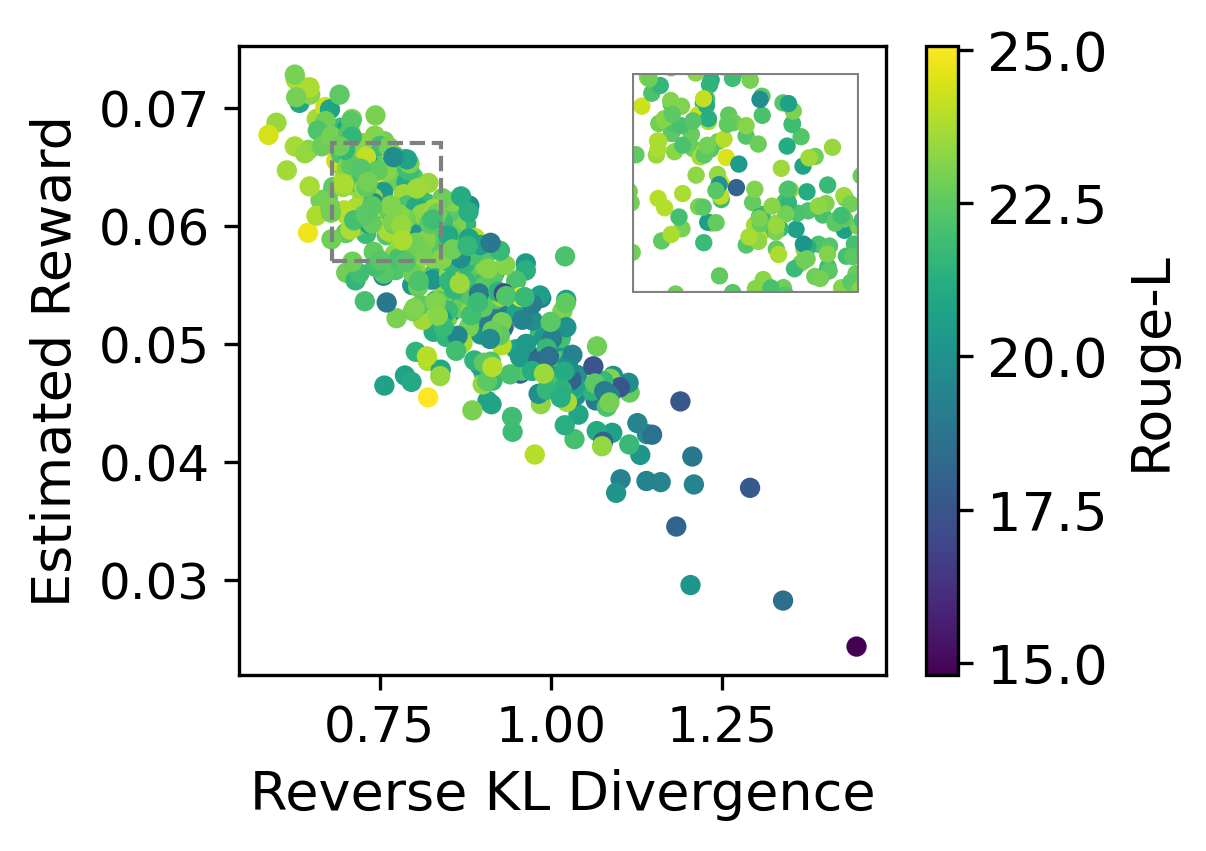} 
\caption{Illustration of the relation of rKLD, implicit reward and Rouge-L. We construct differentiated models by adding random noise to the base model. Lighter colors indicate higher Rouge-L scores.}
\label{fig:toyexperiment}
\end{figure}

In this section, our analysis highlights the inadequacies of rKLD and emphasizes the significance of the reward function introduced in this paper. We conducted an experiment to further illustrate the deficiencies of rKLD discussed in Section \ref{DPKD-method} and the corresponding changes in reward and Rouge-L. As shown in Figure \ref{fig:toyexperiment}, we added random noise to the basic model and simultaneously calculated both the rKLD and the estimated implicit reward presented in this paper during inference. Notably, while the rKLD fluctuates at a relatively low value of 0.001, the Rouge-L score exhibits a fluctuation as high as 6.23.

\begin{figure}[t]
\centering
\includegraphics[width=0.8\columnwidth]{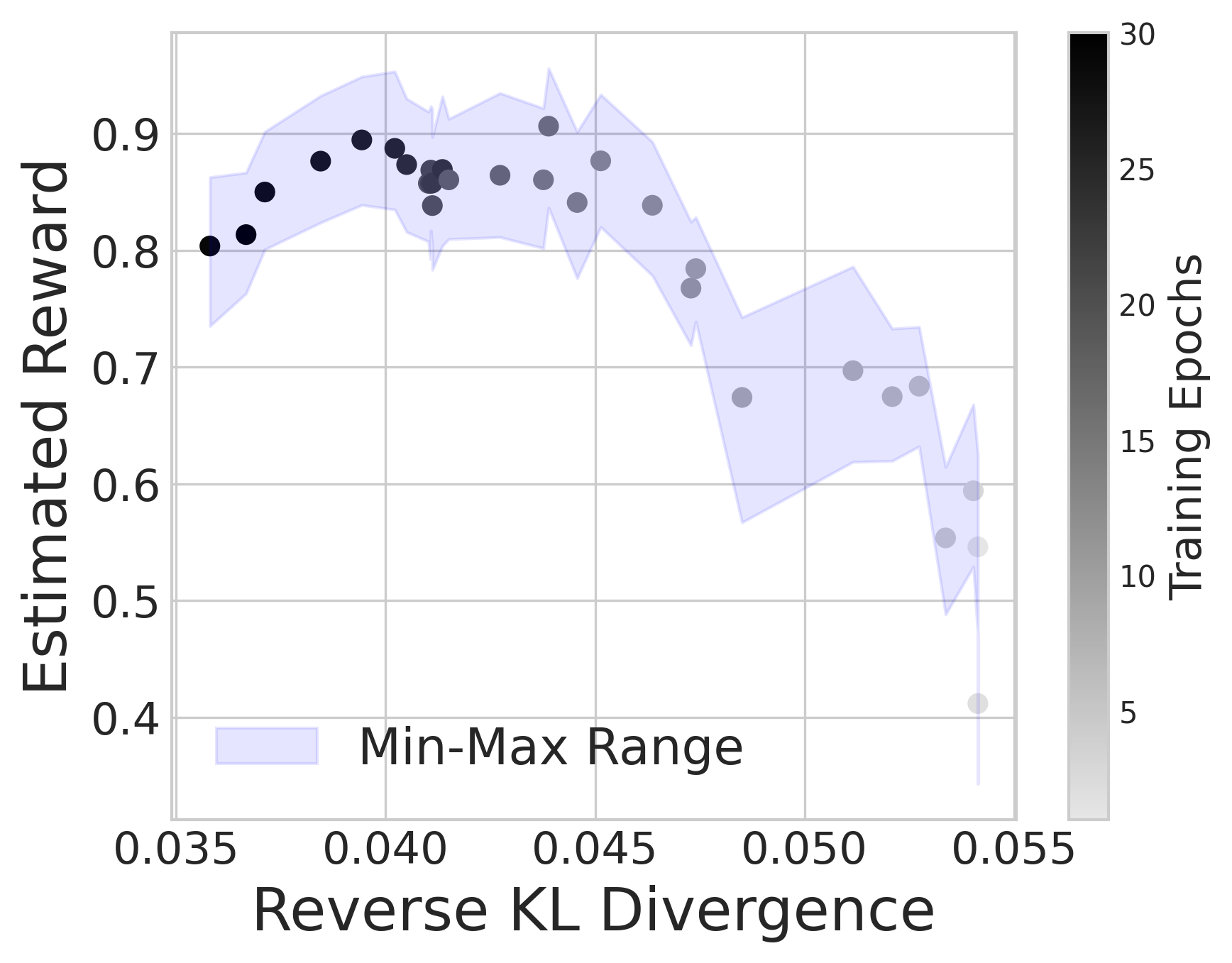} 
\caption{We report the reverse KL divergence, and estimated implicit reward during training. The color of points represents the training epochs. The end of the training falls in the upper left corner, where the KL divergence is low and the reward is high.}
\label{fig:Reward-rKLD-epoch}
\end{figure}

Furthermore, we analyze the implicit reward function discussed in this paper and examine the role of the reward function introduced in KD from both theoretical and experimental perspectives. According to the gradient derivation presented in Section \ref{sec:Gradient Derivation}, the reward function acts as a weight in the gradient update process. When the estimate of the reward function indicates that the current output is biased towards an error, the weight value will increase, resulting in a larger gradient. Conversely, when the model's performance improves and the output distributions of the student and teacher models become similar, the previously mentioned weight term will decrease, leading the model to converge.

We aim to analyze the trend of KL divergence and the reward function of DPKD. To achieve this, we utilize an unbiased approximation of the reward function, which is $ \hat{r}^*(x,y)=\beta\log\frac{q_\theta(y|x)}{p(y|x)}$.
Based on this and the definition of reward estimation, we plot the reward function and KL divergence for the same model (in this case, we use the GPT-2 Base) at different training periods. 

We illustrate the trend of reward during training and the reverse KL divergence distance between the student and teacher models across training epochs in Figure \ref{fig:Reward-rKLD-epoch}.
From the visualization results, we observe that while some model checkpoints exhibit very close KL divergence values to those of the teacher model (all within the range of $0.04\pm0.02$), the differences in reward values still lead to variations in the RougeL scores. From the curve trend in the figure, it can be seen that the optimal point of model training occurs at a left-point arc vertex, while the student model has the lowest KL divergence from the teacher and the reward function is at a higher value. The magnitude of the reward function may be related to the hyperparameter $\beta$ used during training, which adjusts the relative proportions of the KL divergence and the reward function.

\begin{figure}[t]
\centering
  \includegraphics[width=0.7\columnwidth]{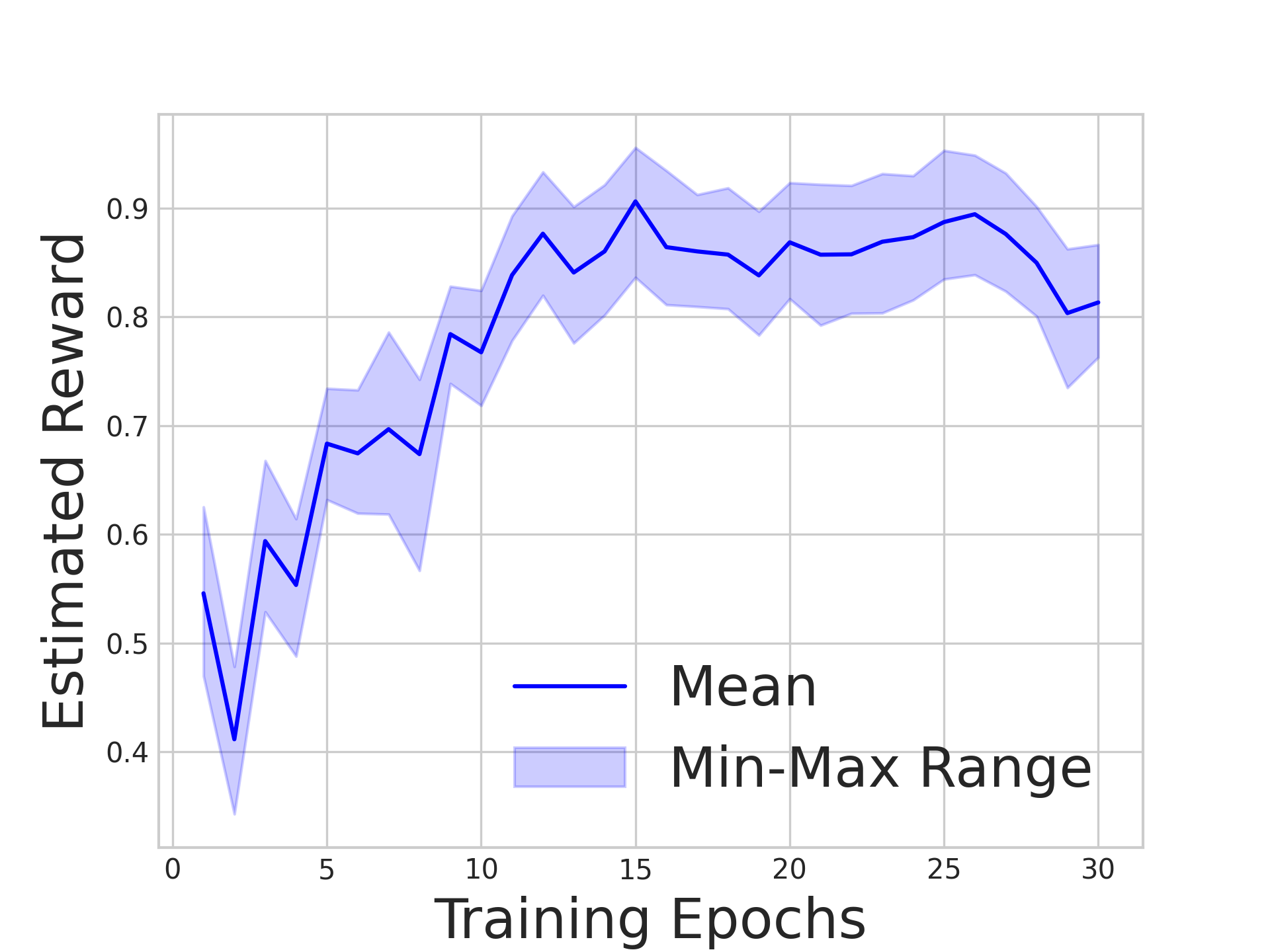} 
  \includegraphics[width=0.465\columnwidth]{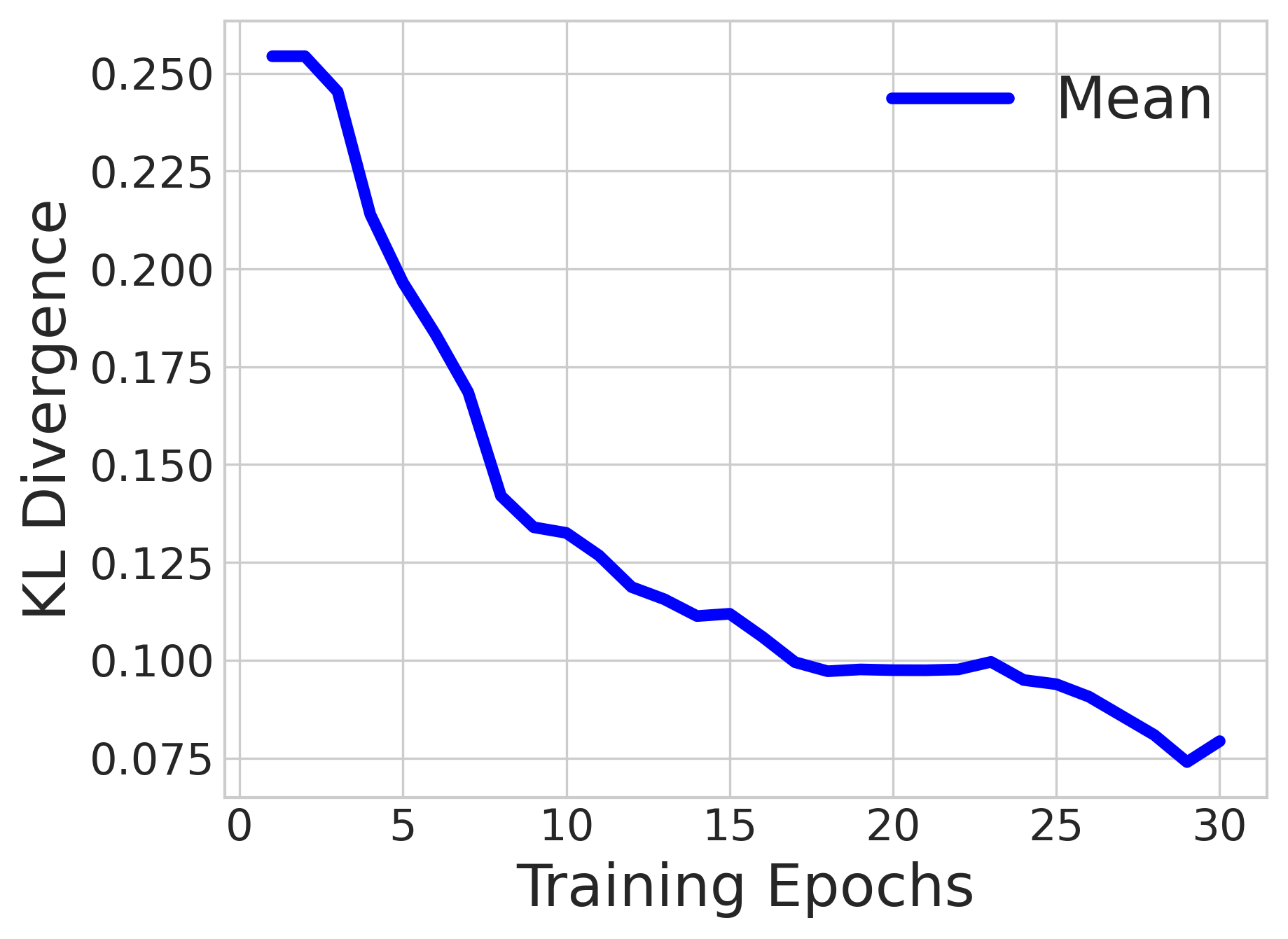} 
  \includegraphics[width=0.465\columnwidth]{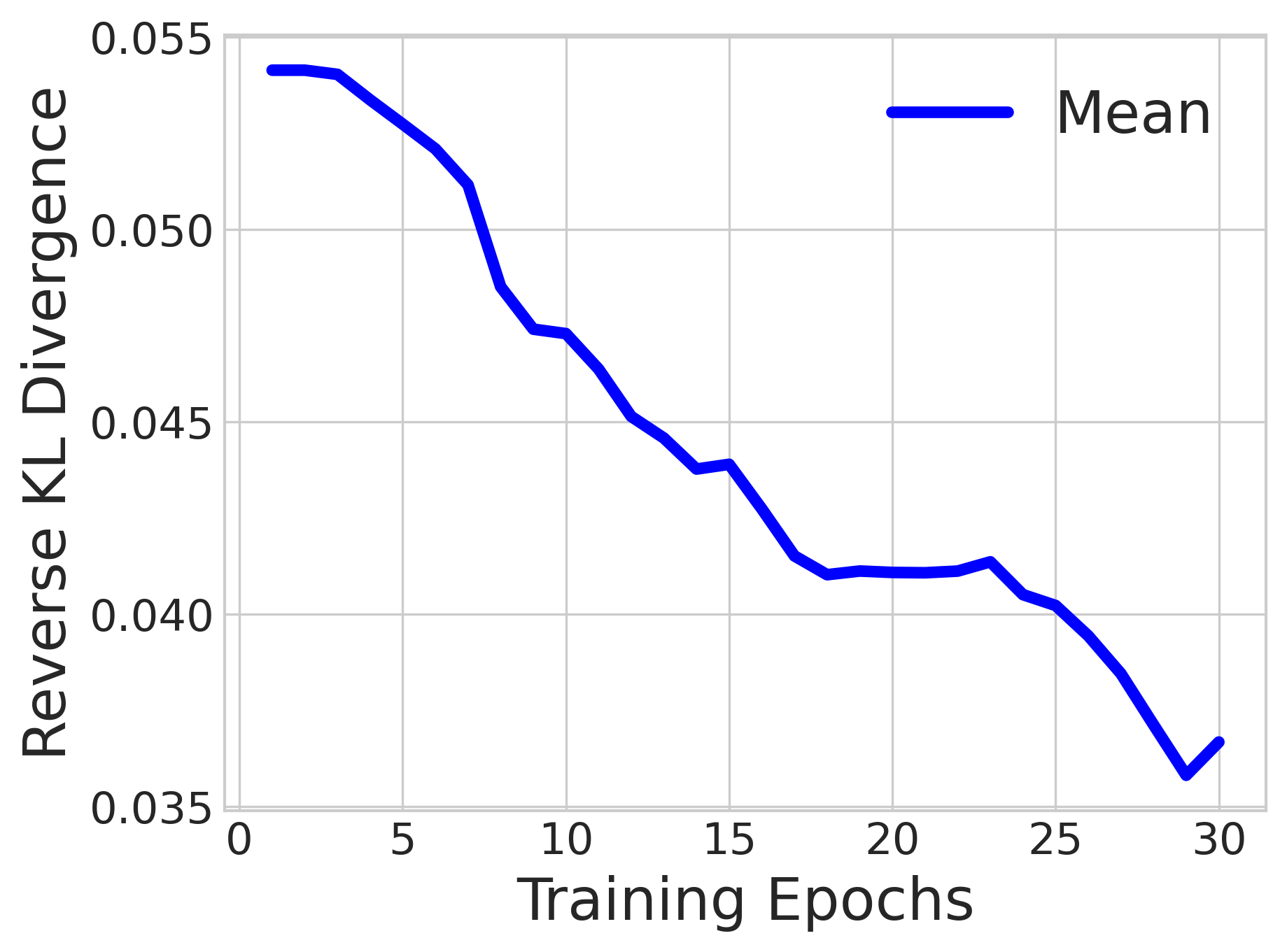} 
  \caption{Reward, KLD, and reverse KLD curves during the distillation process of GPT-2 Base. KLD and reverse KLD show similar trends.}
  \label{fig:reward2}
\end{figure}

Figure \ref{fig:reward2} illustrates the changes in the implicit reward function, KLD, and rKLD during training. We can see that the estimated implicit reward gradually increases within a specific range. The trends of KLD and rKLD are generally similar, but the exact values differ slightly, which is consistent with the conclusions of previous work \cite{Gu2023KnowledgeDO,wu2024rethinking}.

\subsection{Variants of Preference Objective }\label{sec: Deformation of Preference}

In the subject of fine-tuning language models with human feedback, there are many optimization methods based on preference. These various optimization techniques are all based on the original work DPO, and modify aspects such as the preference format and normalization methods. In the context of knowledge distillation for large models, the function form derived in this paper represents a relatively basic form. So similarly, we also conducted preliminary experiments on different methods for calculating preferences, with the aim of inspiring future research. It is important to note that the experiments conducted in this chapter did not perform special parameter fine-tuning.

\begin{table}[t]
\centering
\begin{tabular}{l c}
\hline
\bf Preference Form & \bf RougeL \\ \hline
IPO \cite{IPO} & 18.78 \\
CPO \cite{CPO} & 24.31 \\
SimPO \cite{meng2024simpo} & 24.46 \\ 
DPKD (Ours) & \textbf{ 24.62} \\ \hline
\end{tabular}
\caption{Variants of preference objective. The definitions of each preference are as follows: (1) SimPO = $-\mathbb{E}\log \sigma\left(\frac{\beta}{\left|y_t\right|} \log  q_\theta \left(y_t | x\right)-\frac{\beta}{\left|y_s\right|} \log  q_\theta \left(y_s | x\right)-\gamma\right)$;\\(2) CPO = $-\mathbb{E}\left[\log \sigma \left(\beta \log \frac{q_\theta\left(y_t | x\right)}{q_\theta\left(y_t | x\right)}\right)  -\log  q_\theta \left(y_t | x\right)\right]$;\\(3) IPO = $\mathbb{E}\left(\log \frac{ q_\theta \left(y_t | x\right)}{p\left(y_t | x\right)}-\log \frac{ q_\theta \left(y_s | x\right)}{p\left(y_s | x\right)}-\frac{1}{2 \tau}\right)^2$. }
\label{tab:preference-form}
\end{table}

We experimented with three other forms of preference: IPO, CPO, SimPO. 
Their experiment results and definitions are shown in Table \ref{tab:preference-form}. From the experimental results, we could see that although other forms of preference do not surpass our method, they demonstrate a certain level of effectiveness. \cite{DPOP} illustrated that in the context of learning from preference data, the form of the preference also affects how the relative scores of $y_t$ and $y_s$ grow during the training process. Our findings suggest that these problems and solutions are also applicable to the field of knowledge distillation.

\subsection{Ablation Studies}

\begin{table}[t]\footnotesize
\centering
\begin{tabular}{@{}l ccc@{}}
\hline
\bf Setting & \bf RougeL $\uparrow$  &  \bf rKLD $\downarrow$ &  \bf Reward $\uparrow$ \\ \hline
DPKD           & \bf 24.62  & \bf 0.07  & \bf  0.97 \\
$\,$ w/o LM Loss & 23.47 & 0.24 & 0.93 \\ 
$\,$ w/o Length Norm.    & 22.79   & 0.23 & 0.90 \\ \hline
\end{tabular}
\caption{Ablation studies of DPKD.
``rKLD'' (reverse KL divergence) is the minimum value of the student distance from the teacher model during training. ``Reward'' is the maximum value. LM loss represents the language modeling loss introduced in Section~\ref{sec:Learning Objective} } 
\label{tab: Ablation Studies}
\end{table}


We conduct ablation experiments on the experimental settings and measure the scores on the GPT-2 Base model and DollyEval. We evaluate the results of omitting length normalization and language modeling loss. In Table \ref{tab: Ablation Studies},
KLD and rKLD represent the minimum values of KL divergence and reverse KL divergence from the teacher model during training, respectively. To ensure consistent distribution conditions, the distribution difference is calculated based on the output of the first token. The reward is the maximum value during training and is calculated using the same method described above.
From the results, we can see that the components introduced in the experiment have a positive effect on the results, especially the improvement of implicit reward. The comprehensive method performs well with the teacher model under both reverse KL and forward KL metrics, achieving the highest implicit reward and score.

\subsection{Results of Different Generation Lengths}

\begin{figure}[t] \centering
  \includegraphics[width=0.95\columnwidth]{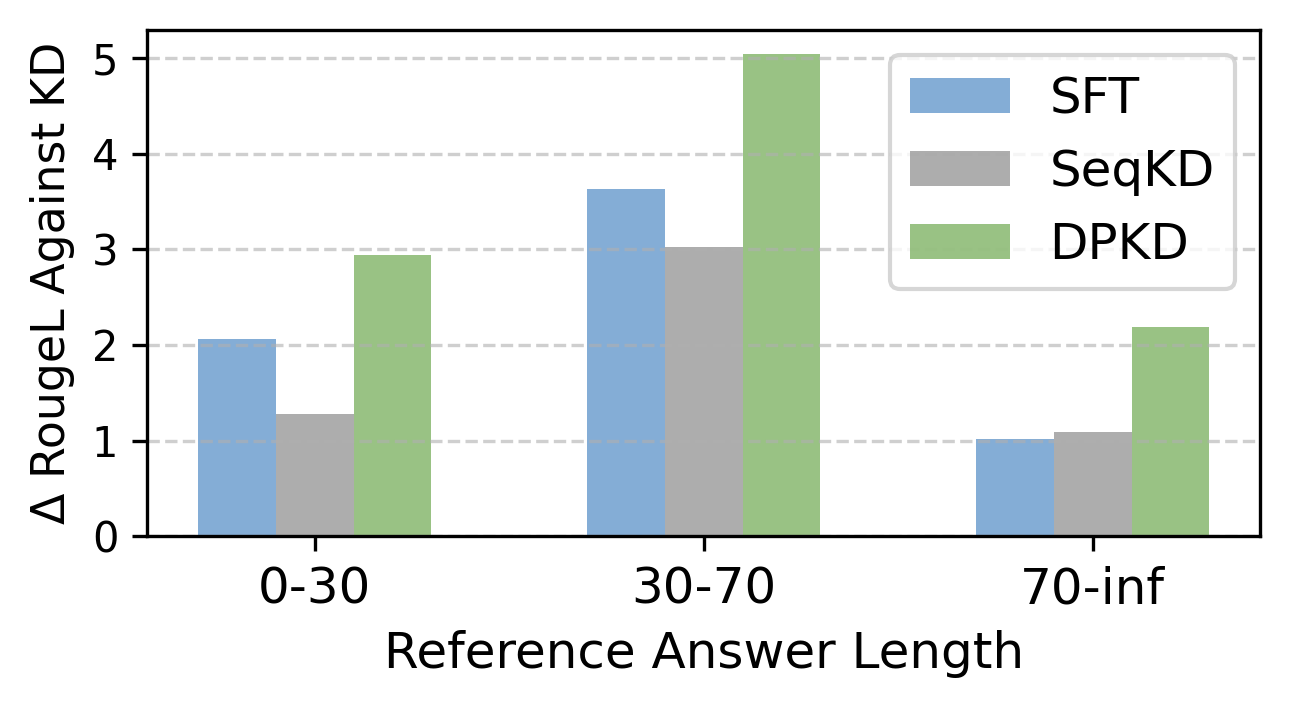} 
  \caption{RougeL score with different ranges of generation lengths. In the case of different ranges of reference label lengths, DPKD scores higher than the baseline. In particular, DPKD stands out when the golden response length is in the middle range. The raw RougeL scores of each method are provided in the Appendix \ref{sec:appendix lenth res}.} 
  \label{fig:different ranges of generation lengths}
\end{figure}

\begin{table}[t]
\centering
\begin{tabular}{l ccc}
\hline
\bf Method & $\left(0, 30\right)$ & $\left(30, 70\right)$ & $\left(70, \infty \right)$ \\ \hline
SeqKD  & 5.15 & 0.00 & 0.00  \\
MiniLLM & 5.30 & 0.93 & \textbf{0.63} \\ 
DPKD & \textbf{ 5.57} & \textbf{1.85} & \textbf{0.63}\\ \hline
\end{tabular}
\caption{ Exact match with different ranges of generation lengths. Our method outperforms the baselines in different ranges of reference lengths, and the results of other baselines tend to be 0 when the length of the golden response is longer.  } 
\label{tab:exact match of length splits}
\end{table}

In this section, we conduct experiments on subsets of the test set with varying response lengths to assess the effectiveness of our approach in scenarios with different output length requirements. We divide the DollyEval test set into three subsets according to the length of the golden response: $\left(0, 30\right)$, $\left(30, 70\right)$ and $\left(70, \infty \right)$, where 30 and 70 are the median and mean of the response lengths, respectively. We then evaluate the response results of various baseline methods and DPKD on these subsets.

The results of the RougeL and exact match are shown in Figure \ref{fig:different ranges of generation lengths} and Table \ref{tab:exact match of length splits}. Our DPKD method performs well on all subsets of response length splits. Notably, DPKD exceeds the baseline by the most in the results with response lengths in the middle. In the exact match results, the scores of many baseline methods drop to 0 when the golden response length is longer, while DPKD still performs well and exceeds the baseline.

\section{Related Work}

\paragraph{Knowledge Distillation (KD)}
is a widely used technique in model training and fine-tuning, and was first proposed in the work \cite{Bucila2006ModelC}, \cite{hinton2015distilling} further enriched and conducted studies on KD. Researchers have explored the application of KD in text generation tasks \cite{song2020lightpaff, zhang2023not, jiao-etal-2020-tinybert, Sun2019PatientKD}. The standard form of KD in NLP is to minimize the KLD between student model and teacher, including train on teacher-generated text \cite{kim-rush-2016-sequence, alpaca} and teacher output at each step \cite{Sanh2019DistilBERTAD}. There are also many works \cite{agarwal2023gkd,passalis2018learning,wen2023f} exploring the metric of distribution distance in KD. Works \cite{huang2022knowledge, cho2019efficacy,mirzadeh2020improved} proposed inadequacy of KLD when the student and teacher model sizes significantly differ. 

\paragraph{KD for LLM}
has received attention in recent years \cite{zhang2022opt, touvron2023llama}. KD in large models is usually divided into black-box distillation \cite{alpaca, vicuna2023} and white-box distillation \cite{kim2024promptkd}. The black-box KD takes the text generated by the teacher as knowledge and performs distillation, while the wite-box KD use the model intermediate layers or output logits distribution \cite{jiao-etal-2020-tinybert, 10.5555/3495724.3496209}. Recent works have recognized the inadequacy of KL divergence in knowledge distillation of LLMs \cite{Gu2023KnowledgeDO} and applied reverse KL divergence to KD of LLMs. Concurrent work \cite{wu2024rethinking} illustrates the shortcomings of reverse KLD and KLD  and designs a KL divergence that adaptively allocates weights.

\section{Conclusion}

In this work we propose a novel method for knowledge distillation of LLMs from the perspective of direct preference learning. We introduce implicit reward and output preference models in knowledge distillation of LLMs and re-formulate the goal of KD. We perform theoretical derivation to obtain a new distillation framework. Experiments show that this framework performs well and is applicable to a wide range of models and data. In addition, we conducts extensive studies on the reward function, preference formula form, showing its importance and providing insights for subsequent work.

\section*{Limitations}

We recognize several limitations of this work. Firstly, the form of preference will affect the situation of model learning $y_t$ and $y_s$. In some cases, direct preference learning will cause the model to focus too much on the relative probability of the two, rather than on the absolute probability of $y_t$ that we expect to obtain. As mentioned in Section \ref{sec: Deformation of Preference}, there are many works that explore and improve different forms of preference. This paper conducts preliminary experiments on these forms of preference and shows that some of the methods may be effective. However, in the field of instruction fine-tuning involved in this paper, the above direct preference situation is not serious. However, the effectiveness of other methods in the preliminary experiments of this paper shows that there may be other problems caused by preferences in the field of LLM KD, and the performance can be further improved through design methods. Section \ref{sec: Deformation of Preference} has introduced preliminary experiments, and we hope that the DPKD methods and this part of the experiment can provide inspiration for future work.



\bibliography{custom}

\begin{thebibliography}{45}
\providecommand{\natexlab}[1]{#1}

\bibitem[{Agarwal et~al.(2023)Agarwal, Vieillard, Stanczyk, Ramos, Geist, and Bachem}]{agarwal2023gkd}
Rishabh Agarwal, Nino Vieillard, Piotr Stanczyk, Sabela Ramos, Matthieu Geist, and Olivier Bachem. 2023.
\newblock Gkd: Generalized knowledge distillation for auto-regressive sequence models.
\newblock \emph{arXiv preprint arXiv:2306.13649}.

\bibitem[{Anil et~al.(2023)Anil, Dai, Firat, Johnson, Lepikhin, Passos, Shakeri, Taropa, Bailey, Chen et~al.}]{anil2023palm}
Rohan Anil, Andrew~M Dai, Orhan Firat, Melvin Johnson, Dmitry Lepikhin, Alexandre Passos, Siamak Shakeri, Emanuel Taropa, Paige Bailey, Zhifeng Chen, et~al. 2023.
\newblock Palm 2 technical report.
\newblock \emph{arXiv preprint arXiv:2305.10403}.

\bibitem[{Azar et~al.(2024)Azar, Guo, Piot, Munos, Rowland, Valko, and Calandriello}]{IPO}
Mohammad~Gheshlaghi Azar, Zhaohan~Daniel Guo, Bilal Piot, Remi Munos, Mark Rowland, Michal Valko, and Daniele Calandriello. 2024.
\newblock A general theoretical paradigm to understand learning from human preferences.
\newblock In \emph{International Conference on Artificial Intelligence and Statistics}, pages 4447--4455. PMLR.

\bibitem[{Bradley and Terry(1952)}]{Bradley1952RankAO}
Ralph~Allan Bradley and Milton~E. Terry. 1952.
\newblock \href {https://api.semanticscholar.org/CorpusID:125209808} {Rank analysis of incomplete block designs: I. the method of paired comparisons}.
\newblock \emph{Biometrika}, 39:324.

\bibitem[{Bucila et~al.(2006)Bucila, Caruana, and Niculescu-Mizil}]{Bucila2006ModelC}
Cristian Bucila, Rich Caruana, and Alexandru Niculescu-Mizil. 2006.
\newblock \href {https://api.semanticscholar.org/CorpusID:11253972} {Model compression}.
\newblock In \emph{Knowledge Discovery and Data Mining}.

\bibitem[{Chiang et~al.(2023)Chiang, Li, Lin, Sheng, Wu, Zhang, Zheng, Zhuang, Zhuang, Gonzalez, Stoica, and Xing}]{vicuna2023}
Wei-Lin Chiang, Zhuohan Li, Zi~Lin, Ying Sheng, Zhanghao Wu, Hao Zhang, Lianmin Zheng, Siyuan Zhuang, Yonghao Zhuang, Joseph~E. Gonzalez, Ion Stoica, and Eric~P. Xing. 2023.
\newblock \href {https://lmsys.org/blog/2023-03-30-vicuna/} {Vicuna: An open-source chatbot impressing gpt-4 with 90\%* chatgpt quality}.

\bibitem[{Cho and Hariharan(2019)}]{cho2019efficacy}
Jang~Hyun Cho and Bharath Hariharan. 2019.
\newblock On the efficacy of knowledge distillation.
\newblock In \emph{Proceedings of the IEEE/CVF international conference on computer vision}, pages 4794--4802.

\bibitem[{Chowdhery et~al.(2023)Chowdhery, Narang, Devlin, Bosma, Mishra, Roberts, Barham, Chung, Sutton, Gehrmann et~al.}]{chowdhery2023palm}
Aakanksha Chowdhery, Sharan Narang, Jacob Devlin, Maarten Bosma, Gaurav Mishra, Adam Roberts, Paul Barham, Hyung~Won Chung, Charles Sutton, Sebastian Gehrmann, et~al. 2023.
\newblock Palm: Scaling language modeling with pathways.
\newblock \emph{Journal of Machine Learning Research}, 24(240):1--113.

\bibitem[{Conover et~al.(2023)Conover, Hayes, Mathur, Xie, Wan, Shah, Ghodsi, Wendell, Zaharia, and Xin}]{DatabricksBlog2023DollyV2}
Mike Conover, Matt Hayes, Ankit Mathur, Jianwei Xie, Jun Wan, Sam Shah, Ali Ghodsi, Patrick Wendell, Matei Zaharia, and Reynold Xin. 2023.
\newblock \href {https://www.databricks.com/blog/2023/04/12/dolly-first-open-commercially-viable-instruction-tuned-llm} {Free dolly: Introducing the world's first truly open instruction-tuned llm}.

\bibitem[{Gu et~al.(2023)Gu, Dong, Wei, and Huang}]{Gu2023KnowledgeDO}
Yuxian Gu, Li~Dong, Furu Wei, and Minlie Huang. 2023.
\newblock \href {https://api.semanticscholar.org/CorpusID:259164722} {Knowledge distillation of large language models}.
\newblock \emph{ArXiv}, abs/2306.08543.

\bibitem[{Hinton et~al.(2015)Hinton, Vinyals, and Dean}]{hinton2015distilling}
Geoffrey Hinton, Oriol Vinyals, and Jeff Dean. 2015.
\newblock Distilling the knowledge in a neural network.
\newblock \emph{arXiv preprint arXiv:1503.02531}.

\bibitem[{Hoffmann et~al.(2022)Hoffmann, Borgeaud, Mensch, Buchatskaya, Cai, Rutherford, Casas, Hendricks, Welbl, Clark et~al.}]{hoffmann2022training}
Jordan Hoffmann, Sebastian Borgeaud, Arthur Mensch, Elena Buchatskaya, Trevor Cai, Eliza Rutherford, Diego de~Las Casas, Lisa~Anne Hendricks, Johannes Welbl, Aidan Clark, et~al. 2022.
\newblock Training compute-optimal large language models.
\newblock \emph{arXiv preprint arXiv:2203.15556}.

\bibitem[{Huang et~al.(2022)Huang, You, Wang, Qian, and Xu}]{huang2022knowledge}
Tao Huang, Shan You, Fei Wang, Chen Qian, and Chang Xu. 2022.
\newblock Knowledge distillation from a stronger teacher.
\newblock \emph{Advances in Neural Information Processing Systems}, 35:33716--33727.

\bibitem[{Jiao et~al.(2020)Jiao, Yin, Shang, Jiang, Chen, Li, Wang, and Liu}]{jiao-etal-2020-tinybert}
Xiaoqi Jiao, Yichun Yin, Lifeng Shang, Xin Jiang, Xiao Chen, Linlin Li, Fang Wang, and Qun Liu. 2020.
\newblock \href {https://doi.org/10.18653/v1/2020.findings-emnlp.372} {{T}iny{BERT}: Distilling {BERT} for natural language understanding}.
\newblock In \emph{Findings of the Association for Computational Linguistics: EMNLP 2020}, pages 4163--4174, Online. Association for Computational Linguistics.

\bibitem[{Kaplan et~al.(2020)Kaplan, McCandlish, Henighan, Brown, Chess, Child, Gray, Radford, Wu, and Amodei}]{kaplan2020scaling}
Jared Kaplan, Sam McCandlish, Tom Henighan, Tom~B Brown, Benjamin Chess, Rewon Child, Scott Gray, Alec Radford, Jeffrey Wu, and Dario Amodei. 2020.
\newblock Scaling laws for neural language models.
\newblock \emph{arXiv preprint arXiv:2001.08361}.

\bibitem[{Kim et~al.(2024)Kim, Jang, and Yang}]{kim2024promptkd}
Gyeongman Kim, Doohyuk Jang, and Eunho Yang. 2024.
\newblock Promptkd: Distilling student-friendly knowledge for generative language models via prompt tuning.
\newblock \emph{arXiv preprint arXiv:2402.12842}.

\bibitem[{Kim and Rush(2016)}]{kim-rush-2016-sequence}
Yoon Kim and Alexander~M. Rush. 2016.
\newblock \href {https://doi.org/10.18653/v1/D16-1139} {Sequence-level knowledge distillation}.
\newblock In \emph{Proceedings of the 2016 Conference on Empirical Methods in Natural Language Processing}, pages 1317--1327, Austin, Texas. Association for Computational Linguistics.

\bibitem[{Lin(2004)}]{lin-2004-rouge}
Chin-Yew Lin. 2004.
\newblock \href {https://aclanthology.org/W04-1013} {{ROUGE}: A package for automatic evaluation of summaries}.
\newblock In \emph{Text Summarization Branches Out}, pages 74--81, Barcelona, Spain. Association for Computational Linguistics.

\bibitem[{Manning and Schutze(1999)}]{manning1999foundations}
Christopher Manning and Hinrich Schutze. 1999.
\newblock \emph{Foundations of statistical natural language processing}.
\newblock MIT press.

\bibitem[{Meng et~al.(2024)Meng, Xia, and Chen}]{meng2024simpo}
Yu~Meng, Mengzhou Xia, and Danqi Chen. 2024.
\newblock Simpo: Simple preference optimization with a reference-free reward.
\newblock \emph{arXiv preprint arXiv:2405.14734}.

\bibitem[{Mirzadeh et~al.(2020)Mirzadeh, Farajtabar, Li, Levine, Matsukawa, and Ghasemzadeh}]{mirzadeh2020improved}
Seyed~Iman Mirzadeh, Mehrdad Farajtabar, Ang Li, Nir Levine, Akihiro Matsukawa, and Hassan Ghasemzadeh. 2020.
\newblock Improved knowledge distillation via teacher assistant.
\newblock In \emph{Proceedings of the AAAI conference on artificial intelligence}, volume~34, pages 5191--5198.

\bibitem[{Nielsen(2021)}]{nielsen2021variational}
Frank Nielsen. 2021.
\newblock On a variational definition for the jensen-shannon symmetrization of distances based on the information radius.
\newblock \emph{Entropy}, 23(4):464.

\bibitem[{Ouyang et~al.(2022)Ouyang, Wu, Jiang, Almeida, Wainwright, Mishkin, Zhang, Agarwal, Slama, Ray et~al.}]{ouyang2022training}
Long Ouyang, Jeffrey Wu, Xu~Jiang, Diogo Almeida, Carroll Wainwright, Pamela Mishkin, Chong Zhang, Sandhini Agarwal, Katarina Slama, Alex Ray, et~al. 2022.
\newblock Training language models to follow instructions with human feedback.
\newblock \emph{Advances in neural information processing systems}, 35:27730--27744.

\bibitem[{Pal et~al.(2024)Pal, Karkhanis, Dooley, Roberts, Naidu, and White}]{DPOP}
Arka Pal, Deep Karkhanis, Samuel Dooley, Manley Roberts, Siddartha Naidu, and Colin White. 2024.
\newblock Smaug: Fixing failure modes of preference optimisation with dpo-positive.
\newblock \emph{arXiv preprint arXiv:2402.13228}.

\bibitem[{Passalis and Tefas(2018)}]{passalis2018learning}
Nikolaos Passalis and Anastasios Tefas. 2018.
\newblock Learning deep representations with probabilistic knowledge transfer.
\newblock In \emph{Proceedings of the European Conference on Computer Vision (ECCV)}, pages 268--284.

\bibitem[{Radford et~al.(2019)Radford, Wu, Child, Luan, Amodei, and Sutskever}]{Radford2019LanguageMA}
Alec Radford, Jeff Wu, Rewon Child, David Luan, Dario Amodei, and Ilya Sutskever. 2019.
\newblock \href {https://api.semanticscholar.org/CorpusID:160025533} {Language models are unsupervised multitask learners}.

\bibitem[{Rafailov et~al.(2024{\natexlab{a}})Rafailov, Hejna, Park, and Finn}]{rafailov2024r}
Rafael Rafailov, Joey Hejna, Ryan Park, and Chelsea Finn. 2024{\natexlab{a}}.
\newblock From $ r $ to $ q $: Your language model is secretly a q-function.
\newblock \emph{arXiv preprint arXiv:2404.12358}.

\bibitem[{Rafailov et~al.(2024{\natexlab{b}})Rafailov, Sharma, Mitchell, Manning, Ermon, and Finn}]{rafailov2024direct}
Rafael Rafailov, Archit Sharma, Eric Mitchell, Christopher~D Manning, Stefano Ermon, and Chelsea Finn. 2024{\natexlab{b}}.
\newblock Direct preference optimization: Your language model is secretly a reward model.
\newblock \emph{Advances in Neural Information Processing Systems}, 36.

\bibitem[{Sanh et~al.(2019)Sanh, Debut, Chaumond, and Wolf}]{Sanh2019DistilBERTAD}
Victor Sanh, Lysandre Debut, Julien Chaumond, and Thomas Wolf. 2019.
\newblock \href {https://api.semanticscholar.org/CorpusID:203626972} {Distilbert, a distilled version of bert: smaller, faster, cheaper and lighter}.
\newblock \emph{ArXiv}, abs/1910.01108.

\bibitem[{Son et~al.(2021)Son, Na, Choi, and Hwang}]{son2021densely}
Wonchul Son, Jaemin Na, Junyong Choi, and Wonjun Hwang. 2021.
\newblock Densely guided knowledge distillation using multiple teacher assistants.
\newblock In \emph{Proceedings of the IEEE/CVF International Conference on Computer Vision}, pages 9395--9404.

\bibitem[{Song et~al.(2020)Song, Sun, Tan, Qin, Lu, Liu, and Liu}]{song2020lightpaff}
Kaitao Song, Hao Sun, Xu~Tan, Tao Qin, Jianfeng Lu, Hongzhi Liu, and Tie-Yan Liu. 2020.
\newblock Lightpaff: A two-stage distillation framework for pre-training and fine-tuning.
\newblock \emph{arXiv preprint arXiv:2004.12817}.

\bibitem[{Sun et~al.(2019)Sun, Cheng, Gan, and Liu}]{Sun2019PatientKD}
Siqi Sun, Yu~Cheng, Zhe Gan, and Jingjing Liu. 2019.
\newblock \href {https://api.semanticscholar.org/CorpusID:201670719} {Patient knowledge distillation for bert model compression}.
\newblock In \emph{Conference on Empirical Methods in Natural Language Processing}.

\bibitem[{Taori et~al.(2023)Taori, Gulrajani, Zhang, Dubois, Li, Guestrin, Liang, and Hashimoto}]{alpaca}
Rohan Taori, Ishaan Gulrajani, Tianyi Zhang, Yann Dubois, Xuechen Li, Carlos Guestrin, Percy Liang, and Tatsunori~B. Hashimoto. 2023.
\newblock Stanford alpaca: An instruction-following llama model.
\newblock \url{https://github.com/tatsu-lab/stanford_alpaca}.

\bibitem[{Touvron et~al.(2023)Touvron, Lavril, Izacard, Martinet, Lachaux, Lacroix, Rozi{\`e}re, Goyal, Hambro, Azhar et~al.}]{touvron2023llama}
Hugo Touvron, Thibaut Lavril, Gautier Izacard, Xavier Martinet, Marie-Anne Lachaux, Timoth{\'e}e Lacroix, Baptiste Rozi{\`e}re, Naman Goyal, Eric Hambro, Faisal Azhar, et~al. 2023.
\newblock Llama: Open and efficient foundation language models.
\newblock \emph{arXiv preprint arXiv:2302.13971}.

\bibitem[{Wang et~al.(2020)Wang, Wei, Dong, Bao, Yang, and Zhou}]{10.5555/3495724.3496209}
Wenhui Wang, Furu Wei, Li~Dong, Hangbo Bao, Nan Yang, and Ming Zhou. 2020.
\newblock Minilm: deep self-attention distillation for task-agnostic compression of pre-trained transformers.
\newblock In \emph{Proceedings of the 34th International Conference on Neural Information Processing Systems}, NIPS '20, Red Hook, NY, USA. Curran Associates Inc.

\bibitem[{Wang et~al.(2023)Wang, Kordi, Mishra, Liu, Smith, Khashabi, and Hajishirzi}]{wang-etal-2023-self-instruct}
Yizhong Wang, Yeganeh Kordi, Swaroop Mishra, Alisa Liu, Noah~A. Smith, Daniel Khashabi, and Hannaneh Hajishirzi. 2023.
\newblock \href {https://doi.org/10.18653/v1/2023.acl-long.754} {Self-instruct: Aligning language models with self-generated instructions}.
\newblock In \emph{Proceedings of the 61st Annual Meeting of the Association for Computational Linguistics (Volume 1: Long Papers)}, pages 13484--13508, Toronto, Canada. Association for Computational Linguistics.

\bibitem[{Wang et~al.(2022)Wang, Mishra, Alipoormolabashi, Kordi, Mirzaei, Naik, Ashok, Dhanasekaran, Arunkumar, Stap, Pathak, Karamanolakis, Lai, Purohit, Mondal, Anderson, Kuznia, Doshi, Pal, Patel, Moradshahi, Parmar, Purohit, Varshney, Kaza, Verma, Puri, Karia, Doshi, Sampat, Mishra, Reddy~A, Patro, Dixit, and Shen}]{wang-etal-2022-super}
Yizhong Wang, Swaroop Mishra, Pegah Alipoormolabashi, Yeganeh Kordi, Amirreza Mirzaei, Atharva Naik, Arjun Ashok, Arut~Selvan Dhanasekaran, Anjana Arunkumar, David Stap, Eshaan Pathak, Giannis Karamanolakis, Haizhi Lai, Ishan Purohit, Ishani Mondal, Jacob Anderson, Kirby Kuznia, Krima Doshi, Kuntal~Kumar Pal, Maitreya Patel, Mehrad Moradshahi, Mihir Parmar, Mirali Purohit, Neeraj Varshney, Phani~Rohitha Kaza, Pulkit Verma, Ravsehaj~Singh Puri, Rushang Karia, Savan Doshi, Shailaja~Keyur Sampat, Siddhartha Mishra, Sujan Reddy~A, Sumanta Patro, Tanay Dixit, and Xudong Shen. 2022.
\newblock \href {https://doi.org/10.18653/v1/2022.emnlp-main.340} {Super-{N}atural{I}nstructions: Generalization via declarative instructions on 1600+ {NLP} tasks}.
\newblock In \emph{Proceedings of the 2022 Conference on Empirical Methods in Natural Language Processing}, pages 5085--5109, Abu Dhabi, United Arab Emirates. Association for Computational Linguistics.

\bibitem[{Wen et~al.(2023)Wen, Li, Du, and Mou}]{wen2023f}
Yuqiao Wen, Zichao Li, Wenyu Du, and Lili Mou. 2023.
\newblock f-divergence minimization for sequence-level knowledge distillation.
\newblock \emph{arXiv preprint arXiv:2307.15190}.

\bibitem[{Wu et~al.(2024)Wu, Tao, Wang, Zhao, and Wong}]{wu2024rethinking}
Taiqiang Wu, Chaofan Tao, Jiahao Wang, Zhe Zhao, and Ngai Wong. 2024.
\newblock Rethinking kullback-leibler divergence in knowledge distillation for large language models.
\newblock \emph{arXiv preprint arXiv:2404.02657}.

\bibitem[{Xu et~al.(2024)Xu, Sharaf, Chen, Tan, Shen, Van~Durme, Murray, and Kim}]{CPO}
Haoran Xu, Amr Sharaf, Yunmo Chen, Weiting Tan, Lingfeng Shen, Benjamin Van~Durme, Kenton Murray, and Young~Jin Kim. 2024.
\newblock Contrastive preference optimization: Pushing the boundaries of llm performance in machine translation.
\newblock \emph{arXiv preprint arXiv:2401.08417}.

\bibitem[{Yuan et~al.(2024)Yuan, Pang, Cho, Sukhbaatar, Xu, and Weston}]{yuan2024self}
Weizhe Yuan, Richard~Yuanzhe Pang, Kyunghyun Cho, Sainbayar Sukhbaatar, Jing Xu, and Jason Weston. 2024.
\newblock Self-rewarding language models.
\newblock \emph{arXiv preprint arXiv:2401.10020}.

\bibitem[{Zhang et~al.(2023)Zhang, Shen, Liu, Liu, Bendersky, Najork, and Zhang}]{zhang2023not}
Rongzhi Zhang, Jiaming Shen, Tianqi Liu, Jialu Liu, Michael Bendersky, Marc Najork, and Chao Zhang. 2023.
\newblock Do not blindly imitate the teacher: Using perturbed loss for knowledge distillation.
\newblock \emph{arXiv preprint arXiv:2305.05010}.

\bibitem[{Zhang et~al.(2022{\natexlab{a}})Zhang, Roller, Goyal, Artetxe, Chen, Chen, Dewan, Diab, Li, Lin et~al.}]{zhang2022opt}
Susan Zhang, Stephen Roller, Naman Goyal, Mikel Artetxe, Moya Chen, Shuohui Chen, Christopher Dewan, Mona Diab, Xian Li, Xi~Victoria Lin, et~al. 2022{\natexlab{a}}.
\newblock Opt: Open pre-trained transformer language models.
\newblock \emph{arXiv preprint arXiv:2205.01068}.

\bibitem[{Zhang et~al.(2022{\natexlab{b}})Zhang, Roller, Goyal, Artetxe, Chen, Chen, Dewan, Diab, Li, Lin, Mihaylov, Ott, Shleifer, Shuster, Simig, Koura, Sridhar, Wang, and Zettlemoyer}]{Zhang2022OPTOP}
Susan Zhang, Stephen Roller, Naman Goyal, Mikel Artetxe, Moya Chen, Shuohui Chen, Christopher Dewan, Mona~T. Diab, Xian Li, Xi~Victoria Lin, Todor Mihaylov, Myle Ott, Sam Shleifer, Kurt Shuster, Daniel Simig, Punit~Singh Koura, Anjali Sridhar, Tianlu Wang, and Luke Zettlemoyer. 2022{\natexlab{b}}.
\newblock \href {https://api.semanticscholar.org/CorpusID:248496292} {Opt: Open pre-trained transformer language models}.
\newblock \emph{ArXiv}, abs/2205.01068.

\bibitem[{Zhou et~al.(2024)Zhou, Liu, Xu, Iyer, Sun, Mao, Ma, Efrat, Yu, Yu et~al.}]{zhou2024lima}
Chunting Zhou, Pengfei Liu, Puxin Xu, Srinivasan Iyer, Jiao Sun, Yuning Mao, Xuezhe Ma, Avia Efrat, Ping Yu, Lili Yu, et~al. 2024.
\newblock Lima: Less is more for alignment.
\newblock \emph{Advances in Neural Information Processing Systems}, 36.

\end{thebibliography}

\appendix

\section{Complete Derivation of DPKD Objective}\label{sec:appendix dpkd objective}

The optimization goal of first step is defined as:

\begin{small}
\begin{equation}
\max_{\theta} \mathbb{E}\,[\,r_p(y|x) - \beta \,  \text{KLD}\,\,(q_{\theta}(y|x)) \| p(y|x) )\,]\label{ap-eq2}
\end{equation}
\end{small}

Transform Equation~\ref{ap-eq2} as follows:

\begin{small}
\begin{equation}
\begin{aligned}
& \max _\theta \mathbb{E} \left[r(x, y)-\beta \log \frac{q_{\theta}(y \mid x)}{p(y \mid x)}\right] \\
 = & \min _\theta \mathbb{E}\left[\log \frac{q_{\theta}(y \mid x)}{p(y \mid x)}-\frac{1}{\beta} r(x, y)\right] \\
 =& \min _\theta \mathbb{E} \left[\log \frac{q_{\theta}(y \mid x)}{\frac{1}{Z(x)} p(y \mid x) \exp \left(\frac{1}{\beta} r(x, y)\right)}-\log Z(x)\right] \\
 =&\min _\theta \mathbb{E}\left[\log \frac{q_{\theta}(y \mid x)}{q^{*}(y \mid x)}-\log Z(x)\right]\label{ap-eq3}
\end{aligned}
\end{equation}
\end{small}

where $q^{*}(y | x) \eqdef \frac{1}{Z(x)} p(y | x) \exp \left(\frac{1}{\beta} r(x, y)\right)$, and $Z\left( x \right)$ is the scaling function of the distribution, which is required to be independent of $\theta$ and  $y$. It is defined as $Z(x) \eqdef \sum_y p(y | x) \exp \left(\frac{1}{\beta} r(x, y)\right)$. 

Substituting into the optimization target, we can get the optimization target:

\begin{small}
\begin{equation}
    \begin{aligned}
 & \min _\theta \mathbb{E}_{x }\left[\mathbb{E}_{y}\left[\log \frac{q_{\theta}(y \mid x)}{q^*(y \mid x)}\right]-\log Z(x)\right] \\
 = & \min _\theta \mathbb{E}_{x}\left[\text{KLD}\left(q_{\theta}(y \mid x) \| q^*(y \mid x)\right)-\log Z(x)\right]
\end{aligned}
\label{ap-eq19}
\end{equation}
\end{small}

Since $Z\left( x \right)$ is a scaling function independent of $\theta$, the minimum value from the optimization objective ~\ref{ap-eq19} is achieved by minimizing the first KL divergence term.
We can derive the optimal solution of Equation~\ref{ap-eq19}:

\begin{small}
\begin{equation}
    q_{\theta}(y \mid x) = q^*(y \mid x) \label{ap-eq4}
\end{equation}
\end{small}
From Equation~\ref{ap-eq4} we can obtain that:

\begin{small}
\begin{equation}
    r^*(x, y)=\beta \log \frac{q^*(y \mid x)}{p(y \mid x)}+\beta \log Z(x) \label{ap-eq5}
\end{equation}
\end{small}

Given the same prompt x, the outputs of the student model and the teacher's model are respectively denoted as $y_s$ and $y_t$. The purpose of KD is to fit the distribution of the student model to teacher model, which can also be understood as we expect the student model to have a greater probability of outputting results similar to the teacher model. From the BT model, we can obtain  the following probability:

\begin{small}
\begin{equation}
\begin{aligned}
    p&^* \left(y_t \succ y_s \mid x\right) \\ 
    & = \sigma\left(\beta \log \frac{q_{\theta}\left(y_t \mid x\right)}{p\left(y_t \mid x\right)}-\beta \log \frac{q_{\theta}\left(y_s \mid x\right)}{p\left(y_s \mid x\right)}\right)\label{ap-3-11}
\end{aligned}
\end{equation}
\end{small}

 Since our goal is to maximize the probability of model outputing $y_t$ rather than $y_s$, which is:

\begin{small}
\begin{equation}
\max _\theta \,\, \mathbb{E} \log \, p^*\left(y_t \succ y_s \mid x\right)
\end{equation}
\end{small}
\begin{small}
\begin{equation}
\min_\theta -\mathbb{E}\left[\log \sigma\left(\beta \log \frac{q_{\theta}\left(y_t \right)}{p\left(y_t \right)}-\beta \log \frac{q_{\theta}\left(y_s \right)}{p\left(y_s \right)}\right)\right] 
\end{equation}
\end{small}

where $q_{\theta}\left(y_t \right)$ denotes $q_{\theta}\left(y_t \mid x\right)$ for short, and same for $p$ and $y_s$.

\section{Complete Derivation of DPKD Gradient }\label{sec:appendix dpkd graadient}

Based on our target definition, we can derive the gradient of DPKD through the basic chain rule and perform analysis. From Equation~\ref{DPKD-loss}, we can perform variable substitution and define $u \eqdef \beta \log \frac{q_\theta\left(y_s \mid x\right)}{p\left(y_s \mid x\right)}-\beta \log \frac{q_\theta\left(y_t \mid x\right)}{p\left(y_t \mid x\right)}$, and the gradient of the loss can be expressed as

\begin{small}
\begin{equation}
\begin{aligned}
   \nabla_\theta \mathcal{L} & = -\nabla_\theta \mathbb{E}_{\left(x, y_t, y_s\right) \sim \mathcal{D}}\left[\log \sigma\left(u\right)\right]\\
   &= -\mathbb{E}_{\left(x, y_t, y_s\right) \sim \mathcal{D}}\left[\frac{\sigma^{\prime}(u)}{\sigma(u)} \nabla_\theta(u)\right]\label{ap-dpkd-loss-dif1}
\end{aligned}
\end{equation}
\end{small}

Define the variable u as follows: 

\begin{small}
\begin{equation}
   u = \beta \log \frac{q_\theta\left(y_s \mid x\right)}{p\left(y_s \mid x\right)}-\beta \log \frac{q_\theta\left(y_t \mid x\right)}{p\left(y_t \mid x\right)}
\end{equation}
\end{small}

Substitute and perform gradient deformation:

\begin{small}
\begin{equation}
   \nabla_\theta \mathcal{L}_{\mathrm{DPKD}}\left(\theta \right) = -\mathbb{E}_{\left(x, y_t, y_s\right) \sim \mathcal{D}}\left[\frac{\sigma^{\prime}(u)}{\sigma(u)} \nabla_\theta(u)\right]\label{ap-dpkd-loss-dif1}
\end{equation}
\end{small}

Since  $\sigma^{\prime}(x)=\sigma(x)(1-\sigma(x))$, we can derive the complete form of gradient:

\begin{small}
\begin{equation}
\begin{aligned}
\nabla_\theta \mathcal{L} =  & -\mathbb{E}\left[\beta  \sigma\left(\beta \log \frac{q_\theta\left(y_t \right)}{p\left(y_t \right)}-\beta \log \frac{q_\theta\left(y_s \right)}{p\left(y_s \right)}\right) \right. \\
& \left. \left[\nabla_\theta \log q_\theta\left(y_t \right)-\nabla_\theta \log q_\theta\left(y_s \right)\right]\right]\label{ap-dpkd-loss-diff-final}
\end{aligned}
\end{equation}
\end{small}

For further analysis, we make a representable approximation of the reward function form in Equation \ref{ap-eq5} 

\begin{small}
    \begin{equation}
    \hat{r}_{\theta}(x, y)=\beta \log \frac{q_{\theta}(y \mid x)}{p(y \mid x)}
\end{equation}
\end{small}

and substitute it into the Equation \ref{ap-dpkd-loss-diff-final}:

\begin{small}
\begin{equation}
\begin{aligned}
\nabla_\theta \mathcal{L} =  & - \beta \, \mathbb{E}\left[  \sigma\left(\hat{r}_{\theta}(x, y_t)-\hat{r}_{\theta}(x, y_s)\right) \right. \\
& \left. \left[\nabla_\theta \log q_\theta\left(y_t \right)-\nabla_\theta \log q_\theta\left(y_s \right)\right]\right]\label{ap-dpkd-loss-diff-final2}
\end{aligned}
\end{equation}
\end{small}

\section{ Complete Result of Different Generation Lengths }\label{sec:appendix lenth res}

\begin{table}[H]\small
\centering
\begin{tabular}{cccc}
\hline
Method & $\left(0, 30\right)$ & $\left(30, 70\right)$ & $\left(70, \infty \right)$ \\ \hline
KD  & 28.04 & 21.65 & 15.07  \\
SeqKD & 29.32 & 24.67 & 16.16 \\ 
SFT & 30.10 & 25.28 & 16.09 \\ 
DPKD &  30.98 & 26.69 & 17.26\\ \hline
\end{tabular}
\caption{ Exact value of RougeL with different ranges of generation lengths.  } 
\label{tab:ap-exact match of lenth splits}
\end{table}

The complete result of Figure \ref{fig:different ranges of generation lengths} is shown in Table \ref{tab:ap-exact match of lenth splits}. From the results, we can see that as the length of the golden response increases, the absolute value of RougeL decreases. This is because the difficulty of the corresponding task of the instruction increases. However, we can still see the advantage of DPKD, as well as the huge lead of DPKD in the middle range of length.


\end{document}